\documentclass[10pt,twocolumn,letterpaper]{article}

\usepackage[pagenumbers]{iccv} 

\usepackage{algorithm}
\usepackage{algorithmic}

\definecolor{iccvblue}{rgb}{0.21,0.49,0.74}
\usepackage[pagebackref,breaklinks,colorlinks,allcolors=iccvblue]{hyperref}

\title{Co-speech Gesture Video Generation via Motion-Based Graph Retrieval}

\author{Yafei Song, Peng Zhang, Bang Zhang\\
Tongyi Lab, Alibaba Group\\
{\tt\small \{huaizhang.syf, futian.zp, zhangbang.zb\}@alibaba-inc.com}
}

\begin{document}
\maketitle

\begin{abstract}
Synthesizing synchronized and natural co-speech gesture videos remains a formidable challenge. Recent approaches have leveraged motion graphs to harness the potential of existing video data. To retrieve an appropriate trajectory from the graph, previous methods either utilize the distance between features extracted from the input audio and those associated with the motions in the graph or embed both the input audio and motion into a shared feature space. However, these techniques may not be optimal due to the many-to-many mapping nature between audio and gestures, which cannot be adequately addressed by one-to-one mapping. To alleviate this limitation, we propose a novel framework that initially employs a diffusion model to generate gesture motions. The diffusion model implicitly learns the joint distribution of audio and motion, enabling the generation of contextually appropriate gestures from input audio sequences. Furthermore, our method extracts both low-level and high-level features from the input audio to enrich the training process of the diffusion model. Subsequently, a meticulously designed motion-based retrieval algorithm is applied to identify the most suitable path within the graph by assessing both global and local similarities in motion. Given that not all nodes in the retrieved path are sequentially continuous, the final step involves seamlessly stitching together these segments to produce a coherent video output. Experimental results substantiate the efficacy of our proposed method, demonstrating a significant improvement over prior approaches in terms of synchronization accuracy and naturalness of generated gestures.
\end{abstract}

\section{Introduction}
\label{sec:intro}

\begin{figure}[htbp]
  \centering
  \includegraphics[width=1.0\linewidth]{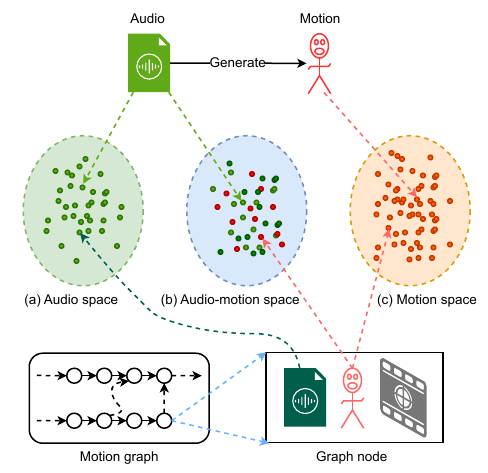}

   \caption{Our key idea is to generate the motion sequence conditioned on the input audio using a diffusion model and then retrieve its nearest trajectory from the pre-constructed motion graph.}
   \label{fig:teaser}
\end{figure}

The synchronization of speech with corresponding co-speech gestures in video content is crucial for enhancing the expressiveness and effectiveness of verbal communication. Generating synchronized and natural-looking co-speech gesture videos remains a complex challenge. Recent advancements have seen various approaches aiming to address this issue, primarily falling into two categories: those utilizing generative models, \eg Generative Adversarial Networks (GANs)~\cite{goodfellow2014gan, mirza2014conditionalGAN, yu2017seqgan, karras2019styleGAN, karras2020styleGAN2, ma2017pose2img, esser2021taming, balakrishnan2018posewarp} or diffusion models~\cite{sohl2015diffusion, ho2020ddpm, songd2021DDIM, dhariwal2021GuidedDiffusion, nichol2021improvedDDPM, rombach2022StableDiffusion, zhang2023controlnet, hu2024animateanyone}, for gesture motion and video generation~\cite{corona2024vlogger, he2024mddiffusion, huang2024make, yang2025showmaker, qian2021sdt}, and those employing motion graphs for retrieval and synthesis~\cite{liu2024tango, zhou2022gvr}.

One category of methodologies focuses on generating gesture motions adopting advanced generative models such as normalizing flow models~\cite{ye2022norm}, Variational AutoEncoder (VAE)~\cite{liu2024ProbTalk, qi2024emotiongesture, ao2022rhythmic, li2021audio2gestures_diversity}, GANs~\cite{habibie2021LS3DCG, qi2024etrans, yoon2020trimodalFMD}, diffusion models~\cite{chen2024diffsheg, mughal2024convofusion, cheng2024siggesture, zhu2023diffgesture, zhang2024motiondiffuse}, and regression models~\cite{ginosar2019s2g, liu2022ha2g, yi2023talkshow, liu2024emage}. These methods are particularly adept at handling the mapping problem between spoken language and gestures. By learning intricate unidirectional mapping or joint distributions between audio and motion, these models can produce contextually appropriate gestures that are nuanced and expressive. However, while these techniques show promise, their application has often been limited to driving 3D models rather than generating real videos. Based on the generated gesture motions, a number of works further drive an image and generate corresponding real videos~\cite{corona2024vlogger, he2024mddiffusion, huang2024make, yang2025showmaker, qian2021sdt, liu2022agnie}. However, due to the limited video generation capability, the results are still not satisfactory.

To obtain more natural video, another prominent approach involves leveraging motion graphs constructed from existing video data. These methods aim to retrieve and synthesize new gesture sequences by exploiting similarities between input audio features and those within the graph. While some studies~\cite{zhou2022gvr} measure the distance in audio space via features extracted from the input audio and the audio from the graph node as shown in \cref{fig:teaser}~(a). Others~\cite{liu2024tango} embed both audio and motion into a shared feature space for retrieval as shown in \cref{fig:teaser}~(b). Despite their effectiveness in utilizing real human motion data, these approaches struggle with the many-to-many mapping problem inherent in translating audio to gesture. The direct use of audio features or embedding audio and motion features into a common space often results in less natural transitions and mismatches due to the complexity of the mappings involved.

To address the limitations of existing methods, we propose a novel framework for Co-speech Gesture Video Generation via Motion-Based Graph Retrieval as shown in \cref{fig:teaser}~(c). Our approach integrates the strengths of diffusion model-based gesture generation with sophisticated motion-based retrieval algorithms. Initially, we employ a diffusion model to generate motion sequences based on input audio, capturing the complex relationships between spoken language and body language. We enhance the training process by incorporating both low-level and high-level features extracted from the input audio, ensuring rich and accurate gesture generation.

Subsequently, we apply a meticulously designed motion-based retrieval algorithm to identify the most suitable path within the motion graph by assessing both global and local similarities in motion. This step ensures that the retrieved segments are not only contextually relevant but also seamlessly integrated. Given the potential discontinuity among nodes in the retrieved path, our final step involves stitching together these segments to produce a coherent and visually appealing video output.

Experimental results demonstrate that our proposed method significantly improves upon previous approaches, offering more natural and accurately synchronized co-speech gesture videos. By combining the advantages of diffusion models and motion graph-based retrieval, we provide a robust solution to the challenges posed by the many-to-many mapping between audio and gestures.

\section{Related Work}

Co-speech gesture generation has gained significant attention, with two primary approaches emerging: direct audio-to-gesture synthesis and motion graph-based methods that retrieve and assemble gestures from existing videos.

\noindent
\textbf{Direct Generation Approaches}

Direct generation methods aim to convert spoken language directly into corresponding gesture animations without relying on pre-existing video datasets. Early efforts in this domain utilized rule-based systems, where predefined rules mapped specific words or phrases to corresponding gestures~\cite{cassell2001beat, huang2012robot}. While these methods were straightforward, they suffered from a lack of flexibility and naturalness, often resulting in robotic and unnatural movements.

More recently, machine learning techniques, particularly deep learning models, have been employed for direct gesture generation. For instance, VAE~\cite{liu2024ProbTalk, qi2024emotiongesture, ao2022rhythmic, li2021audio2gestures_diversity}, GANs~\cite{habibie2021LS3DCG, qi2024etrans, yoon2020trimodalFMD}, regression models~\cite{ginosar2019s2g, liu2022ha2g, yi2023talkshow, liu2024emage}, have been used to model the temporal dependencies between speech and gestures. Although these models improved upon the naturalness and synchronization of generated gestures, they struggled with capturing the full complexity of human movement due to limitations in modeling long-range dependencies and multimodal data.

Diffusion models represent a more recent advancement in this category~\cite{chen2024diffsheg, mughal2024convofusion, cheng2024siggesture, zhu2023diffgesture, zhang2024motiondiffuse}. By implicitly constructing the joint distribution between audio and motion, diffusion models can generate highly contextually appropriate gestures. However, the effectiveness of these models heavily relies on the quality and diversity of training data, as well as the complexity of the feature extraction process. Despite these challenges, diffusion models offer significant improvements in generating nuanced and contextually accurate gestures compared to previous methods.

\noindent
\textbf{Motion Graph-Based Approaches}

Motion graph-based methods utilize pre-recorded human videos to construct a graph structure, enabling retrieval and synthesis of new gesture sequences~\cite{zhou2022gvr, liu2024tango}. Each graph node consists of a piece of audio and one or more continuous video frames. These approaches typically involve creating a motion graph by connecting similar poses between different nodes, allowing for smooth animation generation.

One common strategy involves measuring the distance between features extracted from input audio and those associated with nodes in the graph~\cite{zhou2022gvr}. This method ensures that the retrieved motions are closely aligned with the input audio but may fail to account for the many-to-many mapping between audio and gestures, potentially resulting in less natural or contextually inappropriate gestures.

\begin{figure*}[t]
  \centering
  \includegraphics[width=0.98\linewidth]{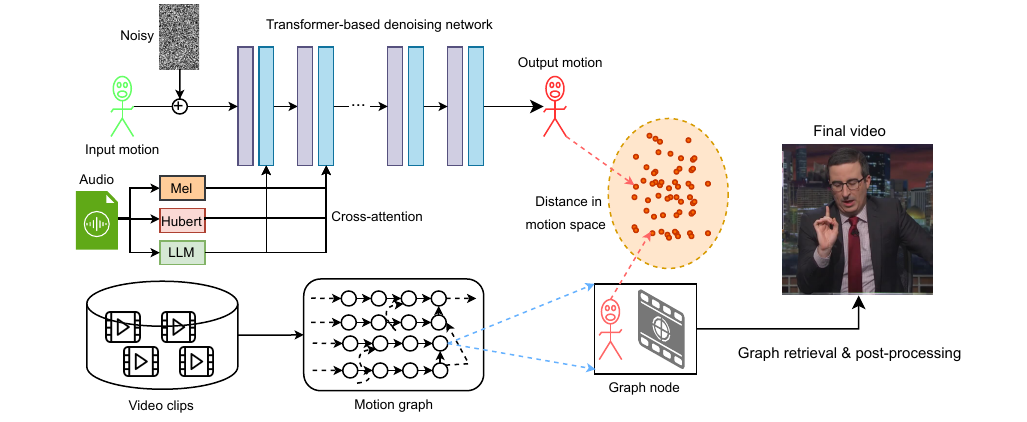}
  \caption{The pipeline of our method. We first train a transformer-based denoising network to generate motion conditioned on the audio, then construct a motion graph using the existing video data. By defining a hybrid motion similarity metric, we could retrieve the optimal trajectory from the motion graph. Combining all nodes in the trajectory, we could get the final video.}
  \label{fig:method}
\end{figure*}

Another approach embeds both the input audio and motion into a shared feature space, facilitating the retrieval of motions that are most similar to the input~\cite{liu2024tango}. While this technique can improve the contextual relevance of the retrieved motions, it also faces challenges related to the accuracy of embeddings and the inherent limitations of one-to-one mappings between audio and gestures.

\section{Method}
\label{sec:method}

In this section, we elaborate on the proposed framework for co-speech gesture video generation via motion-based graph retrieval. As shown in \cref{fig:method}, we first train a diffusion model to learn the joint distribution of audio and gesture. Then, we construct a Motion Graph from a collection of human action videos. At inference time, we generate 3D motion sequences based on input audio using the trained diffusion model and then retrieve matching video segments from the motion graph by aligning generated 3D motions. At last, these retrieved video segments are stitched to form the final coherent video output. Our approach effectively combines the strengths of diffusion models in capturing complex audio-motion relationships with sophisticated motion-based retrieval algorithms, thereby addressing the many-to-many mapping problem between audio and gestures.

\subsection{Generating Gestures Using Diffusion Model}
\label{sec:diffusion}

Our motion generation framework employs a denoising diffusion implicit model (DDIM)~\cite{songd2021DDIM} with a transformer architecture to learn the joint distribution of audio and motion. Given an input audio sequence $A = \{a_t\}_{t=1}^T$, we aim to generate corresponding motion parameters $X = \{x_t\}_{t=1}^T$, where $x_t=\{r\}_{j=1}^J$, $r \in \text{SO}(3)$ is the 3D rotation of a joint, and $J$ denotes the number of joints in SMPL-X~\cite{pavlakos2019smplx}.

\textbf{Data Preprocessing}: Motion parameters are estimated from raw videos using the SMPL-X parametric body model~\cite{cai2023smplerx, yi2023talkshow, cai2023smplerx, pavlakos2019smplx}. For each 3-second video clip (sampled at 30 fps), we obtain motion sequence $X \in r^{90 \times J}$ and corresponding audio features. 

\textbf{Conditional Diffusion Process}: Following the DDIM framework~\cite{songd2021DDIM}, we define the forward process as a Markov chain gradually adding Gaussian noise to the motion data as
\begin{equation}
\begin{split}
q(x_{1:K}|x_0) &:= \prod_{k=1}^K q(x_k|x_{k-1}), \quad q(x_k|x_{k-1}) \\
&:= \mathcal{N}(x_k; \sqrt{\alpha_k}x_{k-1}, (1-\alpha_k)\mathbf{I})
\end{split},
\end{equation}
where $\{\alpha_k\}_{k=1}^K$ defines a monotonically decreasing noise schedule. Our denoising transformer $\epsilon_\theta$ predicts the noise component at each step conditioned on multiple audio features
\begin{equation}
\epsilon_\theta(x_k, k, F) = \text{Transformer}(x_k \oplus E_k \oplus F),
\end{equation}
where $E_k$ is the sinusoidal position encoding for diffusion step $k$, and $F$ represents the fused conditional features.

\textbf{Multi-Modal Conditioning}: We extract three complementary audio features:
1) Mel-Spectrogram features: $F_{\text{mel}} = \text{STFT}(A) \in \mathbb{R}^{T \times 128}$.
2) HuBERT embeddings~\cite{hsu2021hubert}: $F_{\text{hubert}} = \text{HuBERT}(A) \in \mathbb{R}^{T \times 1024}$.
3) LLM (Large Language Model) semantic features: Through automatic speech recognition (ASR)~\cite{radford2023robust} and QWen2-7B~\cite{qwen2}, we obtain text tokens $S = \{s_i\}_{i=1}^M$ with embeddings $F_{\text{token}} \in \mathbb{R}^{M \times 3584}$. These are temporally aligned to motion frames via nearest-neighbor interpolation:
\begin{equation}
F_{\text{LLM}}[t] = F_{\text{token}}[\arg\min_i |t - \frac{T}{M}i|].
\end{equation}
Note that the timestamps obtained by ASR are word-based, not token-based. We suppose each character in one word has the same time duration and get the timestamp of each token via combining all the characters in it.

The final conditioning vector combines these features through adaptive weights
\begin{equation}
F = W_{\text{mel}}F_{\text{mel}} + W_{\text{hubert}}F_{\text{hubert}} + W_{\text{LLM}}F_{\text{LLM}},
\end{equation}
where $\{W_{\cdot}\}$ are learnable projection matrices.

\textbf{Training Objective}: The model is trained via minimizing the noise prediction error
\begin{equation}
\mathcal{L} = \mathbb{E}_{k,x_0,\epsilon}\left[\|\epsilon - \epsilon_\theta(x_k, k, F)\|_2^2\right].
\end{equation}

\textbf{Inference with Temporal Consistency}: For inference on long audio inputs, we follow the in-painting algorithms with diffusion model~\cite{lugmayr2022repaint, chen2024diffsheg} as:
1) Segment input audio into 3-second clips with 0.2-second overlap,
2) Generate initial clip $X^{1:90}$ using ancestral sampling,
3) For subsequent clips $X^{(i)}$, fix the overlapping 6 frames (0.2s) and perform inpainting:
\begin{equation}
\hat{X}^{(i)}_{7:90} = \epsilon_\theta(X^{(i)}_{7:90}, F^{(i)}) \quad \text{s.t.} \quad X^{(i)}_{1:6} = X^{(i-1)}_{85:90},
\end{equation}
4) Blend overlapping regions using linear interpolation.
This approach ensures temporal continuity while allowing error correction through overlapping generations. The complete algorithm is summarized in \cref{alg:diffusion_inference}.

\begin{algorithm}[t]
\caption{Motion Generation with Overlapping Diffusion}
\label{alg:diffusion_inference}
\begin{algorithmic}[1]
\REQUIRE Input audio $A$, trained model $\epsilon_\theta$
\ENSURE Generated motion sequence $X$
\STATE Segment $A$ into $\{A^{(i)}\}$ with 0.2s overlap
\FOR{each clip $A^{(i)}$}
    \IF{$i=1$}
        \STATE Sample $X^{(1)} \sim p_\theta(X|A^{(1)})$
    \ELSE
        \STATE Initialize $X^{(i)}_{1:6} \gets X^{(i-1)}_{85:90}$
        \STATE Inpaint $X^{(i)}_{7:90}$ using $\epsilon_\theta$
    \ENDIF
    \STATE Blend $X^{(i-1)}$ and $X^{(i)}$ in overlap region
\ENDFOR
\RETURN Concatenated motion $X$
\end{algorithmic}
\end{algorithm}

This formulation enables the generation of long, coherent motion sequences while maintaining synchronization with the input audio through multiple complementary conditioning signals. The combination of low-level acoustic features and high-level semantic embeddings allows the model to capture both prosodic and linguistic aspects of gesture generation.

\subsection{Constructing the Motion Graph}
\label{subsec:graph_construction}

The motion graph serves as a structured representation of motion continuity and transition possibilities within the available video data. Our construction process consists of three key phases: node creation, continuous edge establishment, and transition edge identification.

\textbf{Node Representation}: For each video frame $i$, we create a graph node $n_i$ containing the following components:
\begin{itemize}
    \item \textit{Visual Data}: Raw RGB frame $I_i \in \mathbb{R}^{H \times W \times 3}$.
    \item \textit{SMPL-X Parameters}: Body model parameters $\Theta_i = (\beta_i, R_i, t_i)$ where $\beta_i \in \mathbb{R}^{10}$ denotes shape parameters, $R_i \in \text{SO}(3)^{J+1}$ contains joint rotations (including root orientation), $t_i \in \mathbb{R}^3$ represents root translation.
    \item \textit{Joint Positions}: Camera-space coordinates $P_i \in \mathbb{R}^{J \times 3}$ computed via forward kinematics
    \begin{equation}
        P_i = \mathcal{F}_k(\Theta_i, \pi_i),
    \end{equation}
    where $\mathcal{F}_k$ denotes the SMPL-X kinematic function and $\pi_i$ represents camera parameters, \ie focal length $f_i \in \mathbb{R}$.
    \item \textit{Joint Velocities}: Instantaneous velocity $V_i \in \mathbb{R}^{J \times 3}$ calculated through central differencing
    \begin{equation}
        V_i^j = \frac{P_{i+1}^j - P_{i-1}^j}{2\Delta t},
    \end{equation}
    where $\Delta t = 1/\mathrm{fps}$ and $j$ indexes joints.
\end{itemize}

\textbf{Continuous Edges}: We first establish temporal continuity by connecting consecutive frames within original videos:
\begin{equation}
    E_{\text{cont}} = \{(n_i, n_{i+1}) | \forall i < T_v\},
\end{equation}
where $T_v$ denotes the frame count of video $v$. These edges receive a special ``continuous'' flag.

\textbf{Transition Edges}: To enable non-sequential transitions while preserving motion plausibility, we compute potential edges between non-consecutive nodes using a dual-threshold criterion. For any node pair $(n_s, n_t)$ where $s \neq t\pm1$, we calculate \textit{positional discrepancy} for joint $j$:
\begin{equation}
    d_p(s,t,j) = \|P_s^j - P_t^j\|_2.
\end{equation}
and \textit{velocity discrepancy} for joint $j$:
\begin{equation}
    d_v(s,t,j) = \|V_s^j - V_t^j\|_2.
\end{equation}
The adaptive thresholds $\tau_p(s)$ and $\tau_v(s)$ are determined based on local motion characteristics:
\begin{align}
    \tau_p(s) &= \lambda_p \cdot \frac{1}{2}\left(\|P_s - P_{s+1}\|_F + \|P_s - P_{s-1}\|_F\right), \\
    \tau_v(s) &= \lambda_v \cdot \frac{1}{2}\left(\|V_s - V_{s+1}\|_F + \|V_s - V_{s-1}\|_F\right),
\end{align}
where we set $\lambda_p=\lambda_v=1.3$, and $\|\cdot\|_F$ denotes the Frobenius norm across all joints. A transition edge $(n_s, n_t)$ is established if
\begin{equation}
    \frac{1}{J}\sum_{j=1}^J \mathbb{I}\left[d_p(s,t,j) \leq \tau_p(s) \land d_v(s,t,j) \leq \tau_v(s)\right] \geq \mathrm{Th},
\end{equation}
where $\mathbb{I}[\cdot]$ is the indicator function. This ensures that $\mathrm{Th}=95\%$ of joints maintain position and velocity continuity within local motion patterns.

To enable infinite-length video generation without dead-ends, we prune the motion graph to retain only its largest strongly connected component (SCC). With this manner, we identify nodes that form cyclic paths where every node is reachable from all others. The remaining graph preserves original continuous edges while maintaining valid transition edges within the SCC, ensuring any retrieved path can theoretically continue indefinitely by cycling through connected components. The complete motion graph $\mathcal{G} = (\mathcal{V}, \mathcal{E})$ thus contains node set $\mathcal{V} = \{n_i\}_{i=1}^N$ where $N$ is the total node count and edge set $\mathcal{E} = E_{\text{cont}} \cup E_{\text{trans}}$.

This construction methodology ensures that the motion graph preserves original video continuity while enabling plausible transitions between semantically similar motion segments. The adaptive thresholding mechanism accounts for natural motion variability, and the velocity constraints maintain dynamic consistency during transitions. The SCC pruning further guarantees the graph's capacity for infinite generation by eliminating terminal nodes.

\subsection{Retrieving Matching Video Segments}
\label{subsec:retrieval}

Given the generated upper-body motion sequence $X^{\text{gen}} = \{x_t^{\text{gen}}\}_{t=1}^T$ from \cref{alg:diffusion_inference}, our objective is to retrieve the optimal path $\mathcal{P} = (n_1, n_2, ..., n_L)$ through the motion graph $\mathcal{G}$ that best matches $X^{\text{gen}}$. Considering the co-speech gesture generation task, we focus exclusively on upper-body joints, excluding lower-body joints and global body rotations/translations in similarity computations. Our retrieval process consists of two key components: a hybrid motion similarity metric and a pruned tree search algorithm.

\textbf{Motion Similarity Metric}: We propose a composite distance measure combining both rotational and positional discrepancies between generated motions and graph nodes. Let $x^{\text{gen}}_t$ and $n_i$ denote a generated motion frame and a graph node respectively. We compute their distance as:
\begin{equation}
D(x^{\text{gen}}_t, n_i) = \lambda_r D_r(x^{\text{gen}}_t, n_i) + \lambda_p D_p(x^{\text{gen}}_t, n_i),
\end{equation}
where $\lambda_r$ and $\lambda_p$ are balancing weights, with $D_r$ and $D_p$ defined as follows:

1) \textit{Rotational Distance} $D_r$: For each joint $j$ in the upper body ($j \in \mathcal{J}_{\text{upper}}$), we compute the quaternion angular distance between generated rotations $r^{\text{gen}}_{t,j}$ and node rotations $r_{i,j}$:
\begin{equation}
D_r(x^{\text{gen}}_t, n_i) = \frac{1}{|\mathcal{J}_{\text{upper}}|} \sum_{j \in \mathcal{J}_{\text{upper}}} 2\arccos\left(\lvert q^{\text{gen}}_{t,j} \cdot q_{i,j} \rvert \right),
\end{equation}
where $q$ denotes rotation quaternions.

2) \textit{Positional Distance} $D_p$: We compute the Euclidean distance between joint positions derived through forward kinematics:

\begin{equation}
D_p(x^{\text{gen}}_t, n_i) = \frac{1}{|\mathcal{J}_{\text{upper}}|} \sum_{j \in \mathcal{J}_{\text{upper}}} \| \mathcal{F}_k(r^{\text{gen}}_{t})_{j} - P^j_i \|_2,
\end{equation}
where $\mathcal{F}_k(\cdot)$ denotes the forward kinematic function given rotation and $\mathcal{F}_k(\cdot)_j$ denotes the result for joint $j$, and $P^j_i$ is the 3D position of joint $j$ in node $n_i$.

\textbf{Pruned Tree Search Algorithm}: To find the optimal path through the motion graph, we employ a beam search strategy with adaptive pruning as following:

1) \textit{State Representation}: Each search state $s = (n_c, \tau, \mathcal{C})$ consists of:
- Current node $n_c \in \mathcal{V}$
- Time alignment $\tau \in \{1,...,T\}$ indicating correspondence to $x^{\text{gen}}_\tau$
- Accumulated cost $\mathcal{C} = \sum_{t=1}^\tau D(x^{\text{gen}}_t, n_{\pi(t)})$, where $\pi(t)$ maps generated frames to path nodes.

2) \textit{Search Initialization}: Populate the frontier with all nodes having initial cost:
\begin{equation}
\mathcal{C}_0(n_i) = D(x^{\text{gen}}_1, n_i).
\end{equation}

3) \textit{Path Expansion}: For each state $s$ in current frontier:
\begin{itemize}
\item Generate successor states by following all outgoing edges $(n_c, n') \in \mathcal{E}$
\item Update time alignment: $\tau' = \tau + 1$
\item Recode node mapping: $\pi(\tau') = n'$
\item Compute incremental cost: $\Delta\mathcal{C} = D(x^{\text{gen}}_{\tau'}, n')$
\item Update accumulated cost: $\mathcal{C}' = \mathcal{C} + \Delta\mathcal{C} + \beta \cdot \mathbb{I}_{\text{transition}}$, where $\mathbb{I}_{\text{transition}}=0$ for continuous edges and $\mathbb{I}_{\text{transition}}=1$ for transition edges, and $\beta=0.1$ penalizes transition edges to prefer original video continuity.

\end{itemize}

4) \textit{Pruning Strategy}: Maintain only the top-$K$ states (beam width $K=200$) with minimal accumulated cost at each time step $\tau$. Additionally, prune states where:
\begin{equation}
\mathcal{C} > \mathcal{C}_{\text{min}} + \gamma \cdot (\tau/T),
\end{equation}
where $\mathcal{C}_{\text{min}}$ is the current minimum cost and $\gamma=1.5$ controls the pruning threshold.

The complete algorithm terminates when all states reach $\tau = T$, returning the path with minimal final cost $\mathcal{C}_T$. This beam search with adaptive pruning balances exploration of diverse motion possibilities with computational efficiency, ensuring retrieval of contextually appropriate and temporally coherent gesture sequences.

\subsection{Stitching Retrieved Video Segments}
\label{subsec:stitching}

The retrieved path $\mathcal{P}$ from the motion graph may contain transition edges that introduce discontinuities between adjacent nodes. To ensure temporal coherence and visual continuity in the final video output, we implement a two-stage stitching process combining frame interpolation and lip synchronization.

\textbf{Frame Interpolation for Smooth Transitions}: For each transition edge $(n_s, n_t)$ where $n_t$ is not the immediate successor of $n_s$ in the original video, we apply the FILM frame interpolation model~\cite{reda2022film} to generate intermediate frames. Given the two frames before the transition ($I_{s-1}$, $I_s$) and two frames after the transition ($I_t$, $I_{t+1}$), we synthesize two intermediate frames $\{\hat{I}_1, \hat{I}_2\}$ that bridge the motion gap. The interpolation is formulated as:

\begin{equation}
\hat{I}_k = \mathcal{F}_{\text{FILM}}(I_{s-1}, I_{t-1}), \quad k=1,2,
\end{equation}
where $\mathcal{F}_{\text{FILM}}$ denotes the interpolation network.

\textbf{Lip Synchronization}: To enhance audio-visual consistency, we employ the pre-trained Wav2Lip model~\cite{prajwal2020wav2lip} to refine the mouth region of stitched frames. For each interpolated frame $I_k$ aligned with audio segment $A^{(k)}$, we generate lip masks $M_k$ and blend the synthesized mouth regions:

\begin{equation}
I^{\text{final}}_k =\mathcal{F}_{\text{Wav2Lip}}(I_k, A^{(k)}),
\end{equation}
where $\odot$ denotes element-wise multiplication.

Therefore, we improve the temporal consistency to eliminate flickering artifacts, ensuring smooth transitions between stitched frames and lip synchronization.

\section{Experiments}
\label{sec:experiments}

In this section, we introduce the dataset and experimental settings, then compare our results with several previous methods quantitatively and qualitatively. The results demonstrate that the proposed method outperforms these methods with a notable margin. Finally, an ablation study is conducted to verify the effectiveness of several key settings of the proposed method.

\subsection{Dataset and experimental settings}
\label{subsec:dataset}

Following many previous works, such as TANGO~\cite{liu2024tango} and SDT~\cite{qian2021sdt}, we also conduct experiments on the Oliver subset of the SHOW dataset~\cite{yi2023talkshow}, which contains 121 episodes of talk show recordings. The anchor in this show had diverse gestures which is suitable for co-speech gesture generation. Each episode comprises multiple video clips with synchronized speech and upper-body gestures, totaling 28.7 hours of footage. The dataset provides SMPL-X~\cite{pavlakos2019smplx} parameters estimated through elaborate optimization, with joint rotations, global translation and orientation, shape parameters, and camera parameters.

We adopt a stratified splitting strategy to ensure content diversity:
\begin{itemize}
\item Training set: 97 episodes (5,521 clips), which is used to train the diffusion model.
\item Validation set: 12 episodes (757 clips), which is used to validate the training and tune the hyper-parameters.
\item Test set: 12 episodes (833 clips), which is used to test the generation results.
\end{itemize}

For the test episodes, we further partition each episode into:
\begin{itemize}
\item Motion graph construction: 80\% of clips (606 clips), which is used to construct the motion graph. Because the host wears different outfits in each episode, we construct a motion graph for each episode's data.
\item Test queries: 20\% of clips (137 3-10 seconds clips (after removing shorter clips than 3 seconds), which is used for final generation and comparisons. The audio is used as input fore each method, and the video is taken as the ground truth.
\end{itemize}

Our diffusion model uses a 8-layer transformer with 512 hidden dimensions, trained for 3000 epochs using AdamW optimizer (learning rate 2e-4 and reduced on plateau till 1e-5, batch size 512). We extract audio features using:
\begin{itemize}
\item Mel-spectrograms: 128 bins, 33ms window, 33ms hop.
\item HuBERT~\cite{hsu2021hubert}: Large model fine-tuned on LibriSpeech.
\item QWen2-7B~\cite{qwen2}: The features output from the last layer. The model does not output this feature by default, we modify the model to get it. 
\end{itemize}

Motion graph construction employs adaptive thresholds $\lambda_p=1.3$, $\lambda_v=1.3$ with 95\% joint consensus. The retrieval beam search uses $K=200$, $\gamma=1.5$, and $\beta=0.1$. Training is conducted on 4$\times$A100 GPUs with PyTorch 2.0 and testing on one A100.

\begin{table*}[!ht]
\centering
\begin{tabular}{lllll}
\toprule
Method &  CD-FVD$\downarrow$ &FMD$\downarrow$ & Diversity$\uparrow$ & BC$\uparrow$ \\
\midrule
SDT~\cite{qian2021sdt} &  445.1 &38.23 & 1.150 & 0.823 \\
MDDiffusion~\cite{he2024mddiffusion} &  563.6 &40.00 & 1.044 & 0.788 \\
TANGO~\cite{liu2024tango} &  228.3 &29.48 & 1.370 & 0.924 \\
Ours &  170.1 &21.75 & 1.177 & 0.931 \\
\midrule
GT &  0 &0 & 1.340 & 0.941 \\
\bottomrule
\end{tabular}
\caption{Our method achieves the best performance on CD-FVD, FMD, and BC. TANGO exhibits higher diversity, which may stem from its direct reuse of existing motions.}
\label{tab:result}
\end{table*}

\subsection{Quantitative results and comparisons}
\label{subsec:results}

For each test query, we use the trained diffusion model to generate the corresponding motion sequence and take the sequence to retrieve and synthesise the final video based on its corresponding motion graph.

For comparison, we evaluate against three state-of-the-art methods: graph-based method TANGO~\cite{liu2024tango}, diffusion-based method MDDiffusion~\cite{he2024mddiffusion}, and GAN-based method SDT~\cite{qian2021sdt}. For TANGO~\cite{liu2024tango}, we use its Audio-Motion CLIP to retrieve from the motion graph that is identical to ours for a fairer comparison. For MDDiffusion~\cite{he2024mddiffusion} and SDT~\cite{qian2021sdt}, these two methods are designed to drive an image, therefore, we select a clear frame from the corresponding motion graph as the source image.

To quantitatively evaluate the quality of generated videos, we use the content debiased fr\'{e}chet video distance (CD-FVD)~\cite{ge2024cdfvd} instead of FVD~\cite{unterthiner2019fvd}, as FVD is not sensitive to temporal quality. To quantitatively evaluate the generated motion, we use SMPLer-X~\cite{cai2023smplerx} to reconstuct the motion from the generated videos and adopt fr\'{e}chet motion distance (FMD)~\cite{yoon2020trimodalFMD} to measure the motion result. The auto-encoder model used by FMD to compute latent features is trained on the whole Oliver dataset.
To measure the beat aligmented with the audio, we adopt the beat consistency (BC)~\cite{li2021BeatConsitency}. We also report the motion diversity that is described in~\cite{li2021audio2gestures_diversity}.

As presented in \cref{tab:result}, our method achieves the best performance on CD-FVD, FMD, and BC. Compared to the second method, the improvements on CD-FVD (170.1 vs 228.3) and FMD (21.75 vs 29.48) are both remarkable, demonstrating enhanced visual quality and motion fidelity. Our method also achieves a slightly better BC as the diffusion-based motion prior enables better alignment with speech context compared to pure retrieval approaches. While TANGO exhibits higher diversity (1.37 vs 1.177), this may stem from its direct reuse of existing motions, which may preserve dataset biases. Our method maintains competitive beat consistency (0.931 vs 0.924) while generating novel motions. Without a doubt, compared with single-image-based methods~\cite{qian2021sdt, he2024mddiffusion}, our method could achieve obviously better results on all aspects.

\subsection{Ablation studies}
\label{subsec:ablation}

\textit{Impact of LLM Features}: As shown in \cref{tab:ablation}, removing LLM semantic features degrades CD-FVD by 26.5\% and FMD by 31.5\%, validating that high-level semantic cues help generate contextually appropriate gestures. The marginal BC improvement (0.935 vs 0.931) suggests LLM features primarily enhance macro-level motion semantics rather than micro-rhythmic patterns. This result verifies that co-speech gestures not only align to beat but also have a high correlation to the semantical information embedded in the speech.

\textit{Retrieval Strategy}: Replacing our hybrid similarity without rotation distance degrades BC by 9.3\%, while without positional distance, it increases FMD by 14.2\%. This experiment indicates that the combined metric optimally balances local rotation and global spatial constraints, leading to overall better performance.

\begin{table}[!ht]
\centering
\begin{tabular}{lllll}
\toprule
Method &  CD-FVD$\downarrow$ &FMD$\downarrow$ & Div.$\uparrow$ & BC$\uparrow$ \\
\midrule
Ours w/o LLM   & 215.1 &28.60 & 1.124 & 0.935 \\
Ours w/o $D_r$ & 174.4 &22.37 & 1.182 & 0.844 \\
Ours w/o $D_p$ & 178.2 &24.84 & 1.143 & 0.893 \\
Ours           & 170.1 &21.75 & 1.177 & 0.931 \\
\bottomrule
\end{tabular}
\caption{Ablation experiments verify the effectiveness of LLM features, the rotational distance $D_r$ and the position distance $D_p$.}
\label{tab:ablation}
\end{table}

\subsection{Qualitative analysis}
\label{subsec:qualitative}

We present some qualitative result comparisons in \cref{fig:qual} and the videos in the supplement. From the videos, our method produces smoother and higher-fidelity co-speech gesture videos than the compared methods. While TANGO~\cite{liu2024tango} suffers from jitter frames, which may be due to its audio-motion CLIP retrieves more non-continuous segments. MDDiffusion~\cite{he2024mddiffusion} generates blurry and jittery frames, which indicates that current diffusion models have limited ability to generate quick-moving content. SDT~\cite{qian2021sdt} suffers from obvious warping artifacts, which are usually observed in GAN-based methods.

\section{Conclusion}  
This paper presents a novel framework for co-speech gesture video generation that integrates diffusion-based motion synthesis with motion graph retrieval. Our key innovation lies in decoupling the problem into two stages: (1) a diffusion model conditioned on multi-modal audio features (Mel-spectrograms, HuBERT embeddings, and LLM-derived semantics) to generate plausible gesture sequences, and (2) a hybrid motion similarity metric for retrieving and stitching video segments from a pre-constructed motion graph. This approach effectively addresses the many-to-many mapping challenge between speech and gestures while leveraging real motion data for natural video synthesis. Extensive experiments demonstrate state-of-the-art performance in synchronization accuracy and visual quality, outperforming existing methods across multiple metrics, including CD-FVD, FVD, and BeatConsistency.

The main limitations stem from dependency on motion graph construction from video data and occasional transition artifacts. Future work will explore few-shot graph adaptation and enhanced transition modeling. Our framework advances co-speech gesture generation by combining the strengths of learned motion priors and structured motion retrieval, offering practical value for virtual agent animation and human-computer interaction systems.

\begin{figure*}[t]
  \centering
  \includegraphics[width=1.0\linewidth]{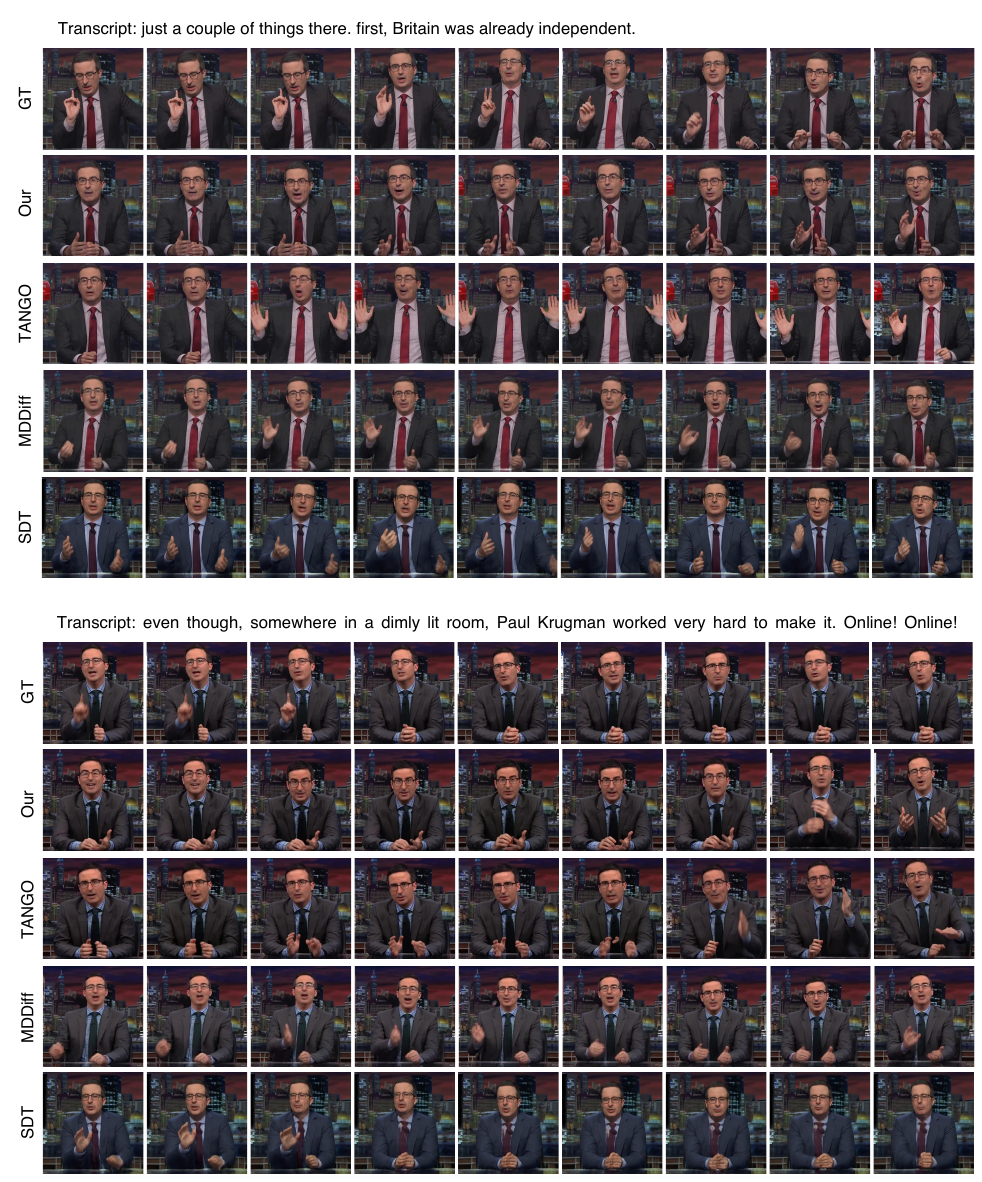}
  \caption{Qualitative comparison showing our method's ability to generate context-specific gestures (\eg, raised arms during emphatic speech). As images hardly represent temporal information, please refer to the videos in the supplement for better observation.}
  \label{fig:qual}
\end{figure*}

{
    \small
    \bibliographystyle{ieeenat_fullname}
    \bibliography{main}

@String(PAMI = {IEEE Trans. Pattern Anal. Mach. Intell.})

@String(CVPR= {IEEE Conf. Comput. Vis. Pattern Recog.})

@String(ICCV= {Int. Conf. Comput. Vis.})

@String(ECCV= {Eur. Conf. Comput. Vis.})

@String(NIPS= {Adv. Neural Inform. Process. Syst.})

@String(TOG= {ACM Trans. Graph.})

@String(TMM  = {IEEE Trans. Multimedia})

@String(ACMMM= {ACM Int. Conf. Multimedia})

@String(ICLR = {Int. Conf. Learn. Represent.})

@String(AAAI = {AAAI})

@String(PAMI  = {IEEE TPAMI})

@String(CVPR  = {CVPR})

@String(ICCV  = {ICCV})

@String(ECCV  = {ECCV})

@String(NIPS  = {NeurIPS})

@String(TOG   = {ACM TOG})

@String(TMM   =	{IEEE TMM})

@String(ACMMM = {ACM MM})

@String(ICLR  = {ICLR})

@article{liu2024tango,
  title={Tango: Co-speech gesture video reenactment with hierarchical audio motion embedding and diffusion interpolation},
  author={Liu, Haiyang and Yang, Xingchao and Akiyama, Tomoya and Huang, Yuantian and Li, Qiaoge and Kuriyama, Shigeru and Taketomi, Takafumi},
  journal={arXiv preprint arXiv:2410.04221},
  year={2024}
}

@article{corona2024vlogger,
  title={Vlogger: Multimodal diffusion for embodied avatar synthesis},
  author={Corona, Enric and Zanfir, Andrei and Bazavan, Eduard Gabriel and Kolotouros, Nikos and Alldieck, Thiemo and Sminchisescu, Cristian},
  journal={arXiv preprint arXiv:2403.08764},
  year={2024}
}

@inproceedings{he2024mddiffusion,
  title={Co-speech gesture video generation via motion-decoupled diffusion model},
  author={He, Xu and Huang, Qiaochu and Zhang, Zhensong and Lin, Zhiwei and Wu, Zhiyong and Yang, Sicheng and Li, Minglei and Chen, Zhiyi and Xu, Songcen and Wu, Xiaofei},
  booktitle=CVPR,
  pages={2263--2273},
  year={2024}
}

@inproceedings{huang2024make,
  title={Make-your-anchor: A diffusion-based 2d avatar generation framework},
  author={Huang, Ziyao and Tang, Fan and Zhang, Yong and Cun, Xiaodong and Cao, Juan and Li, Jintao and Lee, Tong-Yee},
  booktitle=CVPR,
  pages={6997--7006},
  year={2024}
}

@inproceedings{yang2025showmaker,
  title={Showmaker: Creating high-fidelity 2d human video via fine-grained diffusion modeling},
  author={Yang, Quanwei and Guan, Jiazhi and Wang, Kaisiyuan and Yu, Lingyun and Chu, Wenqing and Zhou, Hang and Feng, ZhiQiang and Feng, Haocheng and Ding, Errui and Wang, Jingdong and others},
  booktitle=NIPS,
  volume={37},
  pages={51039--51062},
  year={2024}
}

@inproceedings{qian2021sdt,
  title={Speech drives templates: Co-speech gesture synthesis with learned templates},
  author={Qian, Shenhan and Tu, Zhi and Zhi, Yihao and Liu, Wen and Gao, Shenghua},
  booktitle=ICCV,
  pages={11077--11086},
  year={2021}
}

@inproceedings{zhou2022gvr,
  title={Audio-driven neural gesture reenactment with video motion graphs},
  author={Zhou, Yang and Yang, Jimei and Li, Dingzeyu and Saito, Jun and Aneja, Deepali and Kalogerakis, Evangelos},
  booktitle=CVPR,
  pages={3418--3428},
  year={2022}
}

@inproceedings{chen2024diffsheg,
  title={Diffsheg: A diffusion-based approach for real-time speech-driven holistic 3d expression and gesture generation},
  author={Chen, Junming and Liu, Yunfei and Wang, Jianan and Zeng, Ailing and Li, Yu and Chen, Qifeng},
  booktitle=CVPR,
  pages={7352--7361},
  year={2024}
}

@inproceedings{zhu2023diffgesture,
  title={Taming diffusion models for audio-driven co-speech gesture generation},
  author={Zhu, Lingting and Liu, Xian and Liu, Xuanyu and Qian, Rui and Liu, Ziwei and Yu, Lequan},
  booktitle=CVPR,
  pages={10544--10553},
  year={2023}
}

@article{zhang2024motiondiffuse,
  title={Motiondiffuse: Text-driven human motion generation with diffusion model},
  author={Zhang, Mingyuan and Cai, Zhongang and Pan, Liang and Hong, Fangzhou and Guo, Xinying and Yang, Lei and Liu, Ziwei},
  journal=PAMI,
  volume={46},
  number={6},
  pages={4115--4128},
  year={2024},
  publisher={IEEE}
}

@inproceedings{liu2024emage,
  title={EMAGE: Towards unified holistic co-speech gesture generation via expressive masked audio gesture modeling},
  author={Liu, Haiyang and Zhu, Zihao and Becherini, Giorgio and Peng, Yichen and Su, Mingyang and Zhou, You and Zhe, Xuefei and Iwamoto, Naoya and Zheng, Bo and Black, Michael J},
  booktitle=CVPR,
  pages={1144--1154},
  year={2024}
}

@inproceedings{liu2024ProbTalk,
  title={Towards variable and coordinated holistic co-speech motion generation},
  author={Liu, Yifei and Cao, Qiong and Wen, Yandong and Jiang, Huaiguang and Ding, Changxing},
  booktitle=CVPR,
  pages={1566--1576},
  year={2024}
}

@inproceedings{habibie2021LS3DCG,
  title={Learning speech-driven 3d conversational gestures from video},
  author={Habibie, Ikhsanul and Xu, Weipeng and Mehta, Dushyant and Liu, Lingjie and Seidel, Hans-Peter and Pons-Moll, Gerard and Elgharib, Mohamed and Theobalt, Christian},
  booktitle={Proceedings of the 21st ACM International Conference on Intelligent Virtual Agents},
  pages={101--108},
  year={2021}
}

@inproceedings{ginosar2019s2g,
  title={Learning individual styles of conversational gesture},
  author={Ginosar, Shiry and Bar, Amir and Kohavi, Gefen and Chan, Caroline and Owens, Andrew and Malik, Jitendra},
  booktitle=CVPR,
  pages={3497--3506},
  year={2019}
}

@article{goodfellow2014gan,
  title={Generative adversarial nets},
  author={Goodfellow, Ian and Pouget-Abadie, Jean and Mirza, Mehdi and Xu, Bing and Warde-Farley, David and Ozair, Sherjil and Courville, Aaron and Bengio, Yoshua},
  journal=NIPS,
  volume={27},
  year={2014}
}

@article{ho2020ddpm,
  title={Denoising diffusion probabilistic models},
  author={Ho, Jonathan and Jain, Ajay and Abbeel, Pieter},
  journal=NIPS,
  volume={33},
  pages={6840--6851},
  year={2020}
}

@article{mirza2014conditionalGAN,
  title={Conditional generative adversarial nets},
  author={Mirza, Mehdi and Osindero, Simon},
  journal={arXiv preprint arXiv:1411.1784},
  year={2014}
}

@inproceedings{yu2017seqgan,
  title={Seqgan: Sequence generative adversarial nets with policy gradient},
  author={Yu, Lantao and Zhang, Weinan and Wang, Jun and Yu, Yong},
  booktitle=AAAI,
  volume={31},
  number={1},
  year={2017}
}

@inproceedings{karras2019styleGAN,
  title={A style-based generator architecture for generative adversarial networks},
  author={Karras, Tero and Laine, Samuli and Aila, Timo},
  booktitle=CVPR,
  pages={4401--4410},
  year={2019}
}

@article{karras2020styleGAN2,
  title={Training generative adversarial networks with limited data},
  author={Karras, Tero and Aittala, Miika and Hellsten, Janne and Laine, Samuli and Lehtinen, Jaakko and Aila, Timo},
  journal=NIPS,
  volume={33},
  pages={12104--12114},
  year={2020}
}

@article{ma2017pose2img,
  title={Pose guided person image generation},
  author={Ma, Liqian and Jia, Xu and Sun, Qianru and Schiele, Bernt and Tuytelaars, Tinne and Van Gool, Luc},
  journal=NIPS,
  volume={30},
  year={2017}
}

@inproceedings{balakrishnan2018posewarp,
  title={Synthesizing images of humans in unseen poses},
  author={Balakrishnan, Guha and Zhao, Amy and Dalca, Adrian V and Durand, Fredo and Guttag, John},
  booktitle=CVPR,
  pages={8340--8348},
  year={2018}
}

@inproceedings{nichol2021improvedDDPM,
  title={Improved denoising diffusion probabilistic models},
  author={Nichol, Alexander Quinn and Dhariwal, Prafulla},
  booktitle={International conference on machine learning},
  pages={8162--8171},
  year={2021},
  organization={PMLR}
}

@article{dhariwal2021GuidedDiffusion,
  title={Diffusion models beat gans on image synthesis},
  author={Dhariwal, Prafulla and Nichol, Alexander},
  journal={Advances in neural information processing systems},
  volume={34},
  pages={8780--8794},
  year={2021}
}

@inproceedings{songd2021DDIM,
  title={Denoising Diffusion Implicit Models},
  author={Song, Jiaming and Meng, Chenlin and Ermon, Stefano},
  booktitle=ICLR,
  year={2021}
}

@inproceedings{rombach2022StableDiffusion,
  title={High-resolution image synthesis with latent diffusion models},
  author={Rombach, Robin and Blattmann, Andreas and Lorenz, Dominik and Esser, Patrick and Ommer, Bj{\"o}rn},
  booktitle=CVPR,
  pages={10684--10695},
  year={2022}
}

@inproceedings{esser2021taming,
  title={Taming transformers for high-resolution image synthesis},
  author={Esser, Patrick and Rombach, Robin and Ommer, Bjorn},
  booktitle=CVPR,
  pages={12873--12883},
  year={2021}
}

@inproceedings{sohl2015diffusion,
  title={Deep unsupervised learning using nonequilibrium thermodynamics},
  author={Sohl-Dickstein, Jascha and Weiss, Eric and Maheswaranathan, Niru and Ganguli, Surya},
  booktitle={International conference on machine learning},
  pages={2256--2265},
  year={2015},
  organization={pmlr}
}

@inproceedings{zhang2023controlnet,
  title={Adding conditional control to text-to-image diffusion models},
  author={Zhang, Lvmin and Rao, Anyi and Agrawala, Maneesh},
  booktitle=ICCV,
  pages={3836--3847},
  year={2023}
}

@inproceedings{hu2024animateanyone,
  title={Animate anyone: Consistent and controllable image-to-video synthesis for character animation},
  author={Hu, Li},
  booktitle=CVPR,
  pages={8153--8163},
  year={2024}
}

@inproceedings{yi2023talkshow,
  title={Generating holistic 3d human motion from speech},
  author={Yi, Hongwei and Liang, Hualin and Liu, Yifei and Cao, Qiong and Wen, Yandong and Bolkart, Timo and Tao, Dacheng and Black, Michael J},
  booktitle=CVPR,
  pages={469--480},
  year={2023}
}

@inproceedings{qi2024etrans,
  title={Weakly-supervised emotion transition learning for diverse 3d co-speech gesture generation},
  author={Qi, Xingqun and Pan, Jiahao and Li, Peng and Yuan, Ruibin and Chi, Xiaowei and Li, Mengfei and Luo, Wenhan and Xue, Wei and Zhang, Shanghang and Liu, Qifeng and others},
  booktitle=CVPR,
  pages={10424--10434},
  year={2024}
}

@inproceedings{mughal2024convofusion,
  title={Convofusion: Multi-modal conversational diffusion for co-speech gesture synthesis},
  author={Mughal, Muhammad Hamza and Dabral, Rishabh and Habibie, Ikhsanul and Donatelli, Lucia and Habermann, Marc and Theobalt, Christian},
  booktitle=CVPR,
  pages={1388--1398},
  year={2024}
}

@article{qi2024emotiongesture,
  title={Emotiongesture: Audio-driven diverse emotional co-speech 3d gesture generation},
  author={Qi, Xingqun and Liu, Chen and Li, Lincheng and Hou, Jie and Xin, Haoran and Yu, Xin},
  journal=TMM,
  year={2024},
  publisher={IEEE}
}

@article{liu2022agnie,
  title={Audio-driven co-speech gesture video generation},
  author={Liu, Xian and Wu, Qianyi and Zhou, Hang and Du, Yuanqi and Wu, Wayne and Lin, Dahua and Liu, Ziwei},
  journal=NIPS,
  volume={35},
  pages={21386--21399},
  year={2022}
}

@inproceedings{liu2022ha2g,
  title={Learning hierarchical cross-modal association for co-speech gesture generation},
  author={Liu, Xian and Wu, Qianyi and Zhou, Hang and Xu, Yinghao and Qian, Rui and Lin, Xinyi and Zhou, Xiaowei and Wu, Wayne and Dai, Bo and Zhou, Bolei},
  booktitle=CVPR,
  pages={10462--10472},
  year={2022}
}

@inproceedings{cheng2024siggesture,
  title={SIGGesture: Generalized Co-Speech Gesture Synthesis via Semantic Injection with Large-Scale Pre-Training Diffusion Models},
  author={Cheng, Qingrong and Li, Xu and Fu, Xinghui},
  booktitle={SIGGRAPH Asia 2024 Conference Papers},
  pages={1--11},
  year={2024}
}

@inproceedings{cassell2001beat,
  title={Beat: the behavior expression animation toolkit},
  author={Cassell, Justine and Vilhj{\'a}lmsson, Hannes H{\"o}gni and Bickmore, Timothy},
  booktitle={Proceedings of the 28th annual conference on Computer graphics and interactive techniques},
  pages={477--486},
  year={2001}
}

@inproceedings{huang2012robot,
  title={Robot behavior toolkit: generating effective social behaviors for robots},
  author={Huang, Chien-Ming and Mutlu, Bilge},
  booktitle={Proceedings of the seventh annual ACM/IEEE international conference on Human-Robot Interaction},
  pages={25--32},
  year={2012}
}

@inproceedings{ye2022norm,
  title={Audio-driven stylized gesture generation with flow-based model},
  author={Ye, Sheng and Wen, Yu-Hui and Sun, Yanan and He, Ying and Zhang, Ziyang and Wang, Yaoyuan and He, Weihua and Liu, Yong-Jin},
  booktitle=ECCV,
  pages={712--728},
  year={2022},
  organization={Springer}
}

@article{ao2022rhythmic,
  title={Rhythmic gesticulator: Rhythm-aware co-speech gesture synthesis with hierarchical neural embeddings},
  author={Ao, Tenglong and Gao, Qingzhe and Lou, Yuke and Chen, Baoquan and Liu, Libin},
  journal=TOG,
  volume={41},
  number={6},
  pages={1--19},
  year={2022},
  publisher={ACM New York, NY, USA}
}

@inproceedings{ge2024cdfvd,
  title={On the content bias in fr{\'e}chet video distance},
  author={Ge, Songwei and Mahapatra, Aniruddha and Parmar, Gaurav and Zhu, Jun-Yan and Huang, Jia-Bin},
  booktitle=CVPR,
  pages={7277--7288},
  year={2024}
}

@article{yoon2020trimodalFMD,
  title={Speech gesture generation from the trimodal context of text, audio, and speaker identity},
  author={Yoon, Youngwoo and Cha, Bok and Lee, Joo-Haeng and Jang, Minsu and Lee, Jaeyeon and Kim, Jaehong and Lee, Geehyuk},
  journal=TOG,
  volume={39},
  number={6},
  pages={1--16},
  year={2020},
  publisher={ACM New York, NY, USA}
}

@inproceedings{li2021audio2gestures_diversity,
  title={Audio2gestures: Generating diverse gestures from speech audio with conditional variational autoencoders},
  author={Li, Jing and Kang, Di and Pei, Wenjie and Zhe, Xuefei and Zhang, Ying and He, Zhenyu and Bao, Linchao},
  booktitle=ICCV,
  pages={11293--11302},
  year={2021}
}

@inproceedings{li2021BeatConsitency,
  title={Ai choreographer: Music conditioned 3d dance generation with aist++},
  author={Li, Ruilong and Yang, Shan and Ross, David A and Kanazawa, Angjoo},
  booktitle=ICCV,
  pages={13401--13412},
  year={2021}
}

@inproceedings{pavlakos2019smplx,
  title={Expressive body capture: 3d hands, face, and body from a single image},
  author={Pavlakos, Georgios and Choutas, Vasileios and Ghorbani, Nima and Bolkart, Timo and Osman, Ahmed AA and Tzionas, Dimitrios and Black, Michael J},
  booktitle=CVPR,
  pages={10975--10985},
  year={2019}
}

@inproceedings{prajwal2020wav2lip,
  title={A Lip Sync Expert Is All You Need for Speech to Lip Generation In the Wild},
  author={Prajwal, K R and Mukhopadhyay, Rudrabha and Namboodiri, Vinay P. and Jawahar, C.V.},
  booktitle=ACMMM,
  year={2020},
  pages = {484–492},
  numpages = {9},
  location = {Seattle, WA, USA}
}

@article{qwen2,
    title   = {Qwen2 Technical Report}, 
    author  = {An Yang and Baosong Yang and Binyuan Hui and Bo Zheng and Bowen Yu and Chang Zhou and Chengpeng Li and Chengyuan Li and Dayiheng Liu and Fei Huang and Guanting Dong and Haoran Wei and Huan Lin and Jialong Tang and Jialin Wang and Jian Yang and Jianhong Tu and Jianwei Zhang and Jianxin Ma and Jin Xu and Jingren Zhou and Jinze Bai and Jinzheng He and Junyang Lin and Kai Dang and Keming Lu and Keqin Chen and Kexin Yang and Mei Li and Mingfeng Xue and Na Ni and Pei Zhang and Peng Wang and Ru Peng and Rui Men and Ruize Gao and Runji Lin and Shijie Wang and Shuai Bai and Sinan Tan and Tianhang Zhu and Tianhao Li and Tianyu Liu and Wenbin Ge and Xiaodong Deng and Xiaohuan Zhou and Xingzhang Ren and Xinyu Zhang and Xipin Wei and Xuancheng Ren and Yang Fan and Yang Yao and Yichang Zhang and Yu Wan and Yunfei Chu and Yuqiong Liu and Zeyu Cui and Zhenru Zhang and Zhihao Fan},
    journal = {arXiv preprint arXiv:2407.10671},
    year    = {2024}
}

@article{hsu2021hubert,
  author={Hsu, Wei-Ning and Bolte, Benjamin and Tsai, Yao-Hung Hubert and Lakhotia, Kushal and Salakhutdinov, Ruslan and Mohamed, Abdelrahman},
  title={HuBERT: Self-Supervised Speech Representation Learning by Masked Prediction of Hidden Units},
  year={2021},
  issue_date={2021},
  publisher = {IEEE Press},
  volume = {29},
  issn = {2329-9290},
  url = {https://doi.org/10.1109/TASLP.2021.3122291},
  doi = {10.1109/TASLP.2021.3122291},
  journal = {IEEE/ACM Trans. Audio, Speech and Lang. Proc.},
  month = oct,
  pages = {3451–3460},
  numpages = {10}
}

@inproceedings{cai2023smplerx,
  title={{SMPLer-X}: Scaling up expressive human pose and shape estimation},
  author={Cai, Zhongang and Yin, Wanqi and Zeng, Ailing and Wei, Chen and Sun, Qingping and Yanjun, Wang and Pang, Hui En and Mei, Haiyi and Zhang, Mingyuan and Zhang, Lei and Loy, Chen Change and Yang, Lei and Liu, Ziwei},
  booktitle=NIPS,
  year={2023}
}

@inproceedings{reda2022film,
 title = {FILM: Frame Interpolation for Large Motion},
 author = {Fitsum Reda and Janne Kontkanen and Eric Tabellion and Deqing Sun and Caroline Pantofaru and Brian Curless},
 booktitle = ECCV,
 year = {2022}
}

@inproceedings{radford2023robust,
  title={Robust speech recognition via large-scale weak supervision},
  author={Radford, Alec and Kim, Jong Wook and Xu, Tao and Brockman, Greg and McLeavey, Christine and Sutskever, Ilya},
  booktitle={International conference on machine learning},
  pages={28492--28518},
  year={2023},
  organization={PMLR}
}

@inproceedings{lugmayr2022repaint,
  title={Repaint: Inpainting using denoising diffusion probabilistic models},
  author={Lugmayr, Andreas and Danelljan, Martin and Romero, Andres and Yu, Fisher and Timofte, Radu and Van Gool, Luc},
  booktitle=CVPR,
  pages={11461--11471},
  year={2022}
}

@inproceedings{unterthiner2019fvd,
  title={FVD: A new metric for video generation},
  author={Unterthiner, Thomas and Van Steenkiste, Sjoerd and Kurach, Karol and Marinier, Rapha{\"e}l and Michalski, Marcin and Gelly, Sylvain},
  booktitle={ICLR Workshop DeepGenStruct},
  year={2019}
}
}

\end{document}